\documentclass[10pt,twocolumn,letterpaper]{article}

\usepackage[pagenumbers]{cvpr} % To force page numbers, e.g. for an arXiv version

\usepackage{graphicx}
\usepackage{amsmath}
\usepackage{amssymb}
\usepackage{booktabs}
\usepackage{multirow}
\usepackage{mathrsfs}
\usepackage{appendix}
\usepackage{subcaption}
\usepackage{pifont}
\usepackage[table]{xcolor}

\definecolor{cvprblue}{rgb}{0.21,0.49,0.74}
\usepackage[pagebackref,breaklinks,colorlinks,citecolor=cvprblue]{hyperref}

\definecolor{rowunit}{RGB}{128,128,255}
\definecolor{evaunit01green}{RGB}{54,125,189}
\newcommand{\evagreen}[1]{\textcolor{evaunit01green}{#1}}
\newcommand{\dtplus}[1]{\fontsize{6pt}{0.1em}\selectfont (\textbf{\evagreen{#1}})}

\usepackage{tabularx}
\usepackage[capitalize]{cleveref}
\crefname{section}{Sec.}{Secs.}
\Crefname{section}{Section}{Sections}
\Crefname{table}{Table}{Tables}
\crefname{table}{Tab.}{Tabs.}

\newcommand{\yxq}[1]{{\color{green} }}

\def\recon{{\scshape ReCon}}

\usepackage{graphicx}
\usepackage{amsmath}
\usepackage{amssymb}
\usepackage{booktabs}

\usepackage{footnote}
\usepackage{url}
\usepackage[flushleft]{threeparttable}
\usepackage{enumitem}
\usepackage{algorithm}
\usepackage{algorithmic}
\usepackage{listings}
\usepackage{tabularx}
\usepackage {pifont} 
\usepackage{bm}
\usepackage{tablefootnote}
\usepackage{balance}
\usepackage{colortbl}
\usepackage{multirow}
\usepackage{multicol}

\definecolor{cvprblue}{rgb}{0.21,0.49,0.74}
\usepackage[pagebackref,breaklinks,colorlinks,citecolor=cvprblue]{hyperref}

\definecolor{lightgray}{gray}{0.9}
\definecolor{linecolor}{rgb}{0.82, 0.94, 0.75}

\usepackage[capitalize]{cleveref}
\crefname{section}{Sec.}{Secs.}
\Crefname{section}{Section}{Sections}
\Crefname{table}{Table}{Tables}
\crefname{table}{Tab.}{Tabs.}

\definecolor{evaunit01green}{RGB}{82,208,83}
\definecolor{lowred}{RGB}{238,18,137}

\definecolor{lowerred}{RGB}{255,110,180}

\newcommand{\dplus}[1]{\fontsize{6pt}{0.1em}\selectfont (\textbf{\textcolor{lowred}{#1}})}
\newcommand{\ddplus}[1]{\fontsize{6pt}{0.1em}\selectfont (\textbf{\textcolor{lowerred}{#1}})}

\definecolor{defaultcolor}{RGB}{12,127,17}

\def\recon{{\scshape ReCon}}

\definecolor{learnable}{HTML}{d45c43}
\definecolor{frozen}{HTML}{1f78b4}

\def\paperID{****} % *** Enter the Paper ID here
\def\confName{CVPR}
\def\confYear{2024}

\begin{document}

% %%%%%%%%% TITLE - PLEASE UPDATE
\title{Dynamic Adapter Meets Prompt Tuning:\\
Parameter-Efficient Transfer Learning for Point Cloud Analysis}

\author{Xin Zhou$^{*1}$, Dingkang Liang$^{*1}$, Wei Xu$^{1}$, Xingkui Zhu$^{1}$, Yihan Xu$^{1}$, Zhikang Zou$^{2}$,  Xiang Bai$^{\dag1}$\\
        $^{1}$Huazhong University of Science and Technology, {\tt \{xzhou03,dkliang,xbai\}@hust.edu.cn}\\
        $^{2}$ Baidu Inc., China\\
}

\twocolumn[
\maketitle

{%
\vspace{-10.5mm}
\begin{figure}[H]
\hsize=\textwidth
\centering
\includegraphics[width=2.1\linewidth]{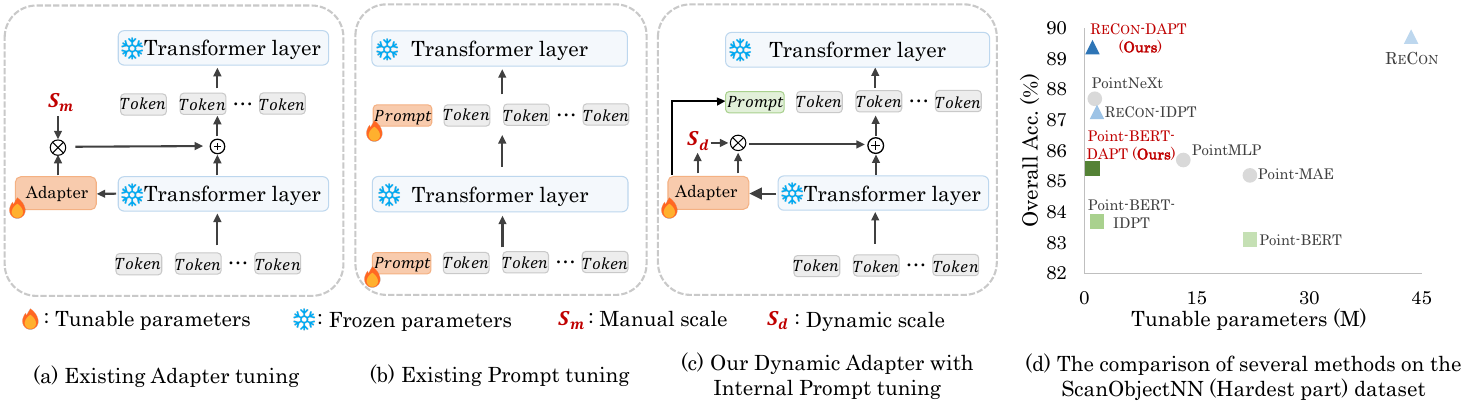}
\vspace{-7mm}
\caption{
(a) Adapter tuning utilizes additional residual blocks with manual scale. (b) Prompt tuning usually introduces extra random initialized prompts into the input space. (c) The proposed Dynamic Adapter generates a dynamic scale for each token and seamlessly integrates with Prompt tuning. (d) The proposed method achieves an ideal trade-off between tunable parameters and performance.}
\vspace{2mm}
\label{fig:intro}
\end{figure}
}]
{\let\thefootnote\relax\footnotetext{* Equal contribution. $\dagger$ Corresponding author.}}

%%%%%%%%% ABSTRACT
\begin{abstract}
\vspace{-5pt}
Point cloud analysis has achieved outstanding performance by transferring point cloud pre-trained models. However, existing methods for model adaptation usually update all model parameters, i.e., full fine-tuning paradigm, which is inefficient as it relies on high computational costs (e.g., training GPU memory) and massive storage space. In this paper, we aim to study parameter-efficient transfer learning for point cloud analysis with an ideal trade-off between task performance and parameter efficiency. To achieve this goal, we freeze the parameters of the default pre-trained models and then propose the Dynamic Adapter, which generates a dynamic scale for each token, considering the token significance to the downstream task. We further seamlessly integrate \textbf{D}ynamic \textbf{A}dapter with \textbf{P}rompt \textbf{T}uning (DAPT) by constructing Internal Prompts, capturing the instance-specific features for interaction. Extensive experiments conducted on five challenging datasets demonstrate that the proposed DAPT achieves superior performance compared to the full fine-tuning counterparts while significantly reducing the trainable parameters and training GPU memory by 95\% and 35\%, respectively. Code is available at \url{https://github.com/LMD0311/DAPT}.
\vspace{-7pt}
\end{abstract}

%%%%%%%%% BODY TEXT
\section{Introduction}
\label{sec:intro}
3D point cloud analysis, a crucial computer vision task, has a broad range of applications, such as autonomous driving~\cite{liu2022epnet++, li2023pillarnext, chen2023voxelnext} and 3D reconstruction~\cite{fan2017point, mittal2022autosdf, chen2023rethinking}. The intrinsic irregularity and sparsity of point clouds pose significant challenges. In the past few years, various deep learning methods~\cite{qi2017pointnet++,cheng2021net, lai2022stratified} have appeared, yielding substantial advancements in this domain. 

To generate robust 3D general representations, using self-supervised learning with vision transformers~\cite{dosovitskiy2020image} to pre-train models is a recent hot topic, such as mask modeling~\cite{he2022masked, pang2022masked} and contrast learning~\cite{qi2023recon,qi2024shapellm}. Further fine-tuning the entire parameters of the pre-trained models on downstream datasets usually leads to faster convergence and significant performance improvements. However, such a fine-tuning strategy is sub-optimal for point cloud analysis, as reflected in the following aspects: 1) Fine-tuning the entire backbone might suffer catastrophic forgetting and break the rich prior knowledge embedded in the pre-training. 2) Fine-tuning for each point cloud analysis dataset requires a separate weights copy, and the storage space overhead may becomes a burden as the number of datasets increases. 3) The computational cost requirements escalate dramatically, especially for larger batch sizes, leading to a substantial increase in GPU memory usage, limiting its accessibility for researchers with weak hardware.

To alleviate these problems, we turn our attention to Parameter-Efﬁcient Transfer Learning (PETL), an alternative fine-tuning strategy that has achieved marked progress in the Natural Language Processing (NLP) domain. Current PETL methods seek to exploit the representational prior in large language models (LLM) by fixing most of the parameters and adjusting only a selected few. Two pioneering PETL paradigms of NLP are Adapter tuning~\cite{houlsby2019parameter,hu2021lora,he2021towards} and Prompt tuning~\cite{lester2021power,li2021prefix}, where the former usually introduces a lightweight network for each Attention/FFN block (Fig.~\ref{fig:intro}(a)) and the latter inserts some external learnable parameters called prompts to the input of each transformer layer (Fig.~\ref{fig:intro}(b)). Both paradigms can obtain performance on par with or surpass the full fine-tuning counterpart while significantly reducing the trainable parameters. 

However, we empirically find that directly using the Adapter/Prompt tuning cannot achieve satisfactory results in point cloud analysis. We argue the main reasons are twofold: \textbf{1)} One key aspect of Adapter tuning is the scale, which is used to adjust the features to match the downstream tasks better, as shown in Fig.~\ref{fig:intro}(a). Existing methods~\cite{he2021towards,chen2022adaptformer,luo2024cheap,jie2023fact} often require manual scale setting as a crucial hyper-parameter, while the value remains constant during inference. This static scale may struggle to adapt to point clouds with complex geometry structures and non-uniform distribution. As a result, such challenges demand a more dynamic and adaptable tuning approach. Simply applying a static scale overlooks the unique characters of each point cloud. \textbf{2)} The Prompt tuning usually adds external random initialized prompts that often lack relevance to the point clouds, as shown in Fig.~\ref{fig:intro}(b). This misalignment makes the network optimization process challenging.

Considering these reasons, in this paper, we propose the \textbf{D}ynamic \textbf{A}dapter and seamlessly integrate it with \textbf{P}rompt \textbf{T}uning, namely `DAPT'. As shown in Fig.~\ref{fig:intro}(c), we introduce an extremely lightweight module to generate a unique scale for each token for Adapter tuning. Such an operator can make sure the Adapter dynamically adjusts each token by considering the significance score and matches the complex 3D samples better. Besides, we propose Internal Prompt tuning, which utilizes the Dynamic Adapter to produce a series of prompts instead of using extra random initialized prompts. Through the Internal Prompt tuning, each prompt will better capture the instance-specific features and be easy to optimize, helping the model capture the global perspective of point clouds.

Extensive experiments on diverse point cloud datasets in different settings demonstrate the effectiveness of our method. In particular, as shown in Fig.~\ref{fig:intro}(d), our DAPT can be simply merged into the state-of-the-art pre-trained models (e.g., Point-BERT~\cite{yu2022point}, \recon~\cite{qi2023recon}) and can significantly reduce the trainable parameters by 95\% and save GPU memory by up to 35\% compared with the full fine-tuning paradigm, while achieving similar or even higher performance, e.g., 2.36\% accuracy improvement on Point-BERT under ScanObjectNN PB\_50\_RS variant.

Our major contributions can be summarized as follows: %\textbf{1)} We reveal the limitation of existing parameter-efficient transfer learning (Adapter tuning and Prompt tuning) of NLP in point cloud analysis. \textbf{2)} We propose the Dynamic Adapter, which generates a unique scale for each token to dynamically adjust features by considering the significance score. \textbf{3)} Based on the Dynamic Adapter, we introduce the Internal Prompt tuning, where we use the Dynamic Adapter to construct the prompt instead of using extra random initialized prompts. 
\begin{itemize}
\item We reveal the limitation of existing parameter-efficient transfer learning (Adapter tuning and Prompt tuning) of NLP in point cloud analysis.
\item We propose the Dynamic Adapter, which generates a unique scale for each token to dynamically adjust features by considering the significance score.
\item Based on the Dynamic Adapter, we introduce the Internal Prompt tuning, where we use the Dynamic Adapter to construct the prompt instead of using extra random initialized prompts.
\end{itemize}

%------------------------------------------------------------------------
\section{Related Work}
\label{sec:relatedwork}

\subsection{Pre-training on Point Cloud}

Transformers have been widely used in 3D vision~\cite{hou2024query,bai2022transfusion,xiong2023cape,zhang2023simple,zhou2022centerformer}, and pre-training on 3D datasets is receiving significant interest. There are two main pretext tasks paradigms for 3D pre-training, including contrastive learning~\cite{xie2020pointcontrast,afham2022crosspoint} and mask modeling~\cite{yu2022point,liu2022masked,liang2024pointmamba}. Contrastive learning-based methods usually contrast between the different views or instances, such as PointContrast~\cite{xie2020pointcontrast} and CrossPoint~\cite{afham2022crosspoint}. Besides, Poursaeed et al.~\cite{poursaeed2020self} defined the estimation of rotation angles as an auxiliary task for contrastive learning. The mask modeling~\cite{zhang2022point,pang2022masked,zha2024towards} typically relies on autoencoders to learn the latent features of the data by reconstructing the original input. For example, Point-BERT~\cite{yu2022point} generates discrete point tokens containing meaningful local information from masked tokens. Point-MAE~\cite{pang2022masked} uses an autoencoder to learn high-level latent features from unmasked patches, aiming to reconstruct the masked point patches. ACT~\cite{dong2022autoencoders} employs an autoencoder as the cross-modal teacher to guide the 3D representation learning. \recon~and \recon ++~\cite{qi2023recon,qi2024shapellm} learns through ensemble distillation from both generative modeling teachers and single/cross-modal contrastive teachers.

The above methods usually fine-tune the entire backbone to adopt the pre-trained model to the downstream 3D tasks effectively, achieving promising performance. However, full fine-tuning is inefficient and may break the well-trained knowledge. In this paper, our focus is on efficiently transferring pre-trained 3D models to downstream tasks.

\subsection{Parameter-Efficient Transfer Learning}

The continuous expansion of pre-trained models demands significant computational resources and consumes considerable storage during fine-tuning. To address these challenges, researchers in the NLP and 2D computer vision domain have explored PETL methods:

Prompt tuning~\cite{lester2021power, jia2022visual} usually adds extra information to the model by introducing latent tokens (prompts) to the task input, enhancing the model's behavior during fine-tuning. Building on prompt tuning, Prefix tuning~\cite{li2021prefix} concatenates tunable prefix vectors to the keys and values of the Attention at every layer. Adapter tuning methods~\cite{houlsby2019parameter, chen2022adaptformer} insert lightweight modules named Adapter into the FFN in a complementary approach, allowing for incremental model refinements. LoRA~\cite{hu2021lora} further adopts a low-rank approximation to update weight matrices in the Attention. Building on these foundations, several variations have been proposed, including varying the placement of Adapters~\cite{he2021towards, lian2022scaling, zhu2021counter}, empowering the network to select the best method combination~\cite{mao2022unipelt}, and implementing strategies to minimize the trainable parameters~\cite{jie2023fact, karimi2021compacter, yin20231, jie2023revisiting}. 

So far, only one method, IDPT~\cite{zha2023instance} focuses on parameter-efficient transfer learning on point cloud analysis. IDPT extends the Prompt tuning with a DGCNN~\cite{wang2019dynamic} to extract instance-aware prompts for model fine-tuning instead of using static prompts. Unlike IDPT, we propose the
Dynamic Adapter and seamlessly integrate it with Prompt Tuning, which significantly reduces the tunable parameters and achieves impressive performance.

\section{Preliminary}
\label{sec:preliminary}
In this section, we revisit the vision transformer and introduce Adapter tuning and Prompt tuning as preliminary.

\subsection{Vision Transformer} 
In a vision transformer (ViT)~\cite{dosovitskiy2020image}, an image is divided into $N$ non-overlapping patches and transferred to a 1D sequence. The sequence is further processed by $L$-layer transformer blocks, along with a classification token. The input $\boldsymbol{x} \in \mathbb{R}^{(N+1)\times d}$, where $d$ is the embedding dimension, is first transformed to queries $\boldsymbol{Q} \in \mathbb{R}^{(N+1)\times d}$, keys $\boldsymbol{K} \in \mathbb{R}^{(N+1)\times d}$ and values $\boldsymbol{V} \in \mathbb{R}^{(N+1)\times d}$. After that, we can calculate the self-attention with an attention layer.

The output of the attention layer will be sent to an FFN module with residual bypasses to extract information in channels. The transformer block can be written as below:
\begin{equation}
\begin{aligned}
\boldsymbol{x}^ {\prime}_i & = \operatorname{Attention}\left (\operatorname{LN}\left (\boldsymbol{x}_{i-1}\right) \right) + \boldsymbol{x}_{i-1},\\
\boldsymbol{x}_i & = \operatorname{FFN}\left (\operatorname{LN}\left (\boldsymbol{x}^ {\prime}_i\right )\right ) + \boldsymbol{x}^ {\prime}_i,
\end{aligned}
\end{equation}
where $\boldsymbol{x}_i$ is the output of $i$-th block, LN is layer norm. Recently, some work~\cite{pang2022masked,yu2022point} has adapted the ViT for point cloud analysis and utilized self-supervision for pre-training. In ViT, the Attention and FFN components contain most parameters. It would be reasonable to fix them for PETL, but doing so would damage performance.

\subsection{Parameter-Efficient Fine-tuning}
\textbf{Adapter tuning}~\cite{houlsby2019parameter,he2021towards, chen2022adaptformer} is a module that acts as a bottleneck and inserts a few parameters into the transformer. It includes a downward projection to decrease the feature dimension, a non-linear function, and an upward projection to project back to the original dimension. Specifically, given the input as $\boldsymbol{x} \in \mathbb{R}^{(N+1)\times d}$, the output is calculated by
\begin{equation}
  \operatorname{Output}=S_m\times\left [ \left ( \phi\left (\boldsymbol{x}W_{d}^T\right )\right )W_{u}^T\right ],
\end{equation}
where $W_{d}\in \mathbb{R}^{r\times d}$, $\phi(\cdot )$, $W_{u}\in \mathbb{R}^{d\times r}$ ($r \ll d$), and $S_m$ represent the down-projection weight, non-linear function, up-projection weight, and scale, respectively. Note that $S_m$ is a key hyper-parameter that needs to be manually set in existing methods~\cite{he2021towards,chen2022adaptformer,luo2024cheap,jie2023fact}.

\textbf{Prompt tuning}~\cite{lester2021power,li2021prefix, jia2022visual} creates random initialized tokens as prompts. These prompts are then incorporated into the input of a transformer block or attention layer, where they interact with the original tokens through self-attention. In the fine-tuning process, the weights of the backbone network remain frozen, and only the prompts are updated. Given classification token $\boldsymbol{T}_{cls} \in \mathbb{R}^{1 \times d}$ and patch tokens $\boldsymbol{T}\in \mathbb{R}^{N\times d}$, we can represent the inserted prompts as $\boldsymbol{P}_{i} =\{\boldsymbol{P}_i^k\in \mathbb{R} ^{1 \times d}\mid k\in \mathbb{N}, 1\le k\le n\}$, where $n$ is the number of prompts. The output of $i$-th layer ($L_{i}(\cdot)$)  $\boldsymbol{x}_i \in \mathbb{R}^{(N+n+1)\times d}$ are obtained by
% \begin{equation}
%   x^ {\prime} = \left [ p;x \right ].
% \end{equation}
\begin{equation}
    \boldsymbol{x}_{i} =L_i \left ( \left [\boldsymbol{T}_{cls}; \boldsymbol{P}_i; \boldsymbol{T} \right ] \right ).
\end{equation}

However, while these tuning strategies achieve promising results in NLP and 2D vision, they lack targeted design for 3D. As shown in Tab.~\ref{tab:tuning_strategy}, when it comes to the hardest (\textit{i.e.}, PB\_T50\_RS) variant of ScanObjectNN~\cite{uy2019revisiting}, Adapter and VPT have an accuracy gap of 1.73\% and 4.09\% when compared to full fine-tuning, respectively. Thus, designing a parameter-efficient fine-tuning method that performs well on 3D data is meaningful.

\begin{figure*}[!ht]
	\begin{center}
		\includegraphics[width=0.99\linewidth]{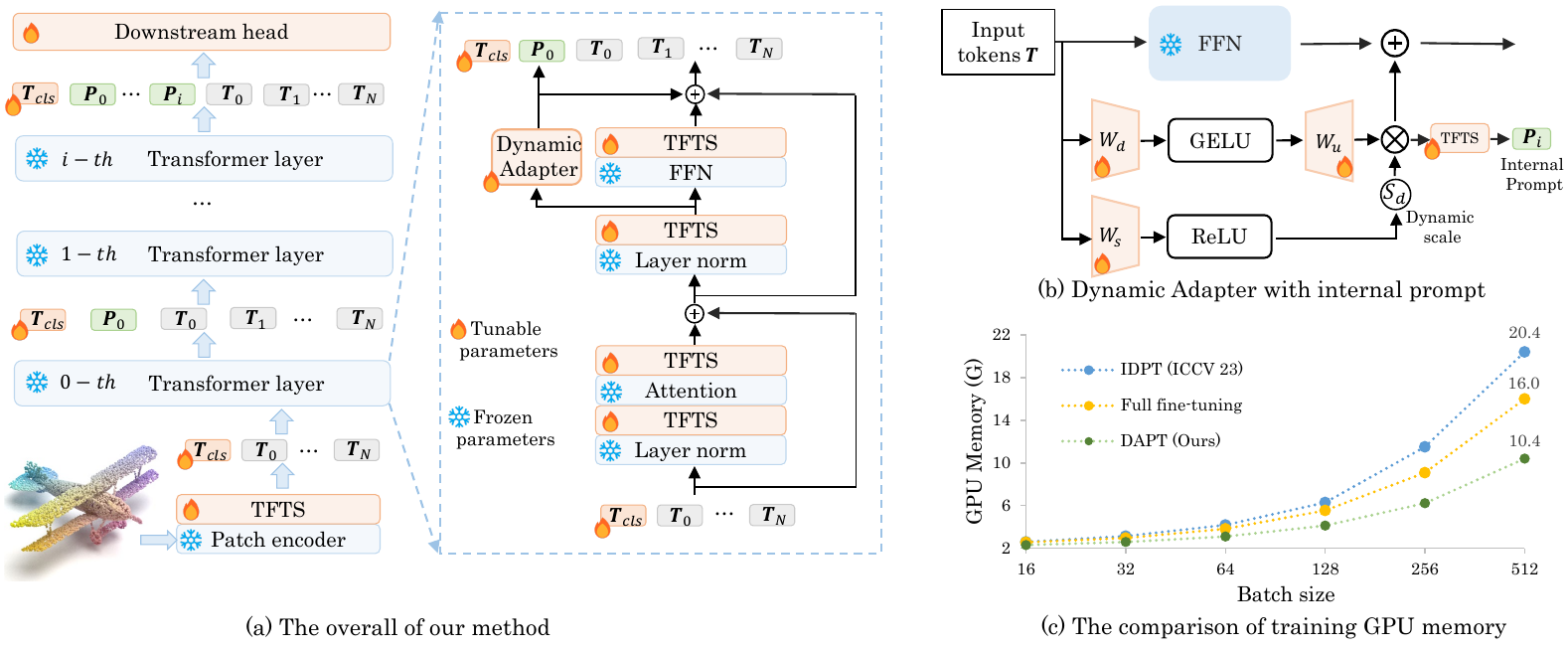}
	\end{center}
 \vspace{-16pt}
	\caption{
	(a) The pipeline of our DAPT. During training, we fix the entire backbone, only fine-tuning the newly added parameters. (b) The detail of the Dynamic Adapter with Internal Prompt. (c) Comparison of GPU memory usage among full fine-tuning, IDPT~\cite{zha2023instance} and ours.}
	\label{fig:pipeline}
 \vspace{-10pt}
\end{figure*}

\begin{table}[!t]
\footnotesize
\setlength{\tabcolsep}{1.6mm}
\centering
\caption{The overall accuracy (\%) for Adapter and Prompt tuning strategies on three variants of ScanObjectNN~\cite{uy2019revisiting} is reported. `\#TP’ means the number of tunable parameters. Linear probing indicates head-tuned only.}
\vspace{-10pt}
\label{tab:tuning_strategy}
\begin{tabular}{ lcccc }
\toprule
Tuning Strategy & \#TP(M) &OBJ\_BG &OBJ\_ONLY & PB\_T50\_RS \\
\midrule
Point-MAE~\cite{pang2022masked} & 22.1 & 90.02 & 88.29 & 85.18 \\
Linear probing & 0.3  & 87.26\dtplus{-2.76} & 84.85\dtplus{-3.44} & 75.99\dtplus{-9.19}\\
\midrule
+ Adapter~\cite{houlsby2019parameter} & 0.9  & 89.50\dtplus{-0.52} & 88.64\dplus{+0.35} & 83.93\dtplus{-1.25}\\
+ VPT~\cite{jia2022visual} & 0.4  & 87.26\dtplus{-2.76} & 87.09\dtplus{-1.20} &81.09\dtplus{-4.09} \\
\bottomrule
\vspace{-15pt}
\end{tabular}
\end{table}

\section{Methodology}
Fig.~\ref{fig:pipeline}(a) illustrates an overview of the proposed DAPT, which consists of three key components: a Task-agnostic Feature Transform Strategy (TFTS) for feature regulation, a Dynamic Adapter for feature modification, and an Internal Prompt selected from the output of Dynamic Adapter. The specifics of our DAPT will be discussed below.

\subsection{Task-agnostic Feature Transform Strategy}
\label{sec:tfts}

3D pre-training models, such as masked autoencoders~\cite{pang2022masked,zhang2022point} or contrastive learning~\cite{afham2022crosspoint,qi2023recon}, have demonstrated their effectiveness in acquiring a generic feature representation space. However, these representations are often task-agnostic, thus requiring fine-tuning for downstream tasks. Hence, we introduce a frustratingly simple Task-agnostic Feature Transform Strategy (TFTS), adding two small learnable factors to transform the model's features~\cite{lian2022scaling}. Given the input $\boldsymbol{x}\in \mathbb{R}^{(N+1)\times d}$, the output $\boldsymbol{y}\in \mathbb{R}^{(N+1)\times d}$ is calculated as follows:
\begin{equation}
  \boldsymbol{y} = \boldsymbol{\gamma} \times \boldsymbol{x} + \boldsymbol{\beta},
\end{equation}
where $\boldsymbol{\gamma},\boldsymbol{\beta}\in\mathbb{R}^{1 \times d}$ are learnable factors. Such a simple strategy effectively converts features to match downstream representations of 3D data, resulting in considerable performance improvements while only introducing an additional 0.5\% parameters to the baselines.

\subsection{Dynamic Adapter}
\label{sec:adapter}
Adapter tuning~\cite{houlsby2019parameter, he2021towards, chen2022adaptformer} from NLP and 2D vision usually involves manual scale, significantly affecting the adapted model's performance. Furthermore, this scale must be adjusted manually as a hyper-parameter for each specific task, which is time-consuming and inconvenient. To address this problem, we proposed the Dynamic Adapter, comprising two projection layers: a downward projection with parameters $\boldsymbol{W}_d \in \mathbb{R}^{r \times d}$ and an upward projection with parameters $\boldsymbol{W}_{u} \in \mathbb{R}^{d \times r}$, as shown in Fig.~\ref{fig:pipeline}(b).

The two projection layers are connected by a GELU function. We adopt a parallel MLP to generate dynamic scale $S_d$ based on variant point cloud features. Specifically, we utilize a scoring weight matrix $\boldsymbol{W}_{s} \in \mathbb{R}^{1 \times d}$ to calculate a scale $S_d$ for each token. With $\boldsymbol{x}$ as the input, our dynamic scale can be expressed as
\begin{equation}
S_d = \operatorname{ReLU}\left ( \boldsymbol{x}\boldsymbol{W}_{s}^T \right ).
\end{equation}
Note that we utilize the ReLU to select the positive scale and set the rest as zero, as we think only significant tokens need to be adjusted, and it should depend on the unique characters of each point cloud. In general, the output of our Dynamic Adapter $\boldsymbol{x}^{\prime}$ can be described as 
\begin{equation}
\boldsymbol{x}^{\prime}=S_d \times \left [ \left ( \operatorname{GELU} \left ( \boldsymbol{x}\boldsymbol{W}_{d}^T \right )\right ) \boldsymbol{W}_{u}^T\right ].
\end{equation}
To mitigate the influence of Adapter outputs during the initial stages of model training, we initialize $\boldsymbol{W}_u$ to zero.

\subsection{Internal Prompt Tuning}
\label{sec:prompt}
In this subsection, we introduce how to combine our Dynamic Adapter with Prompt tuning. The existing Prompt tuning methods~\cite{lester2021power,li2021prefix,jia2022visual} usually adopt external random initialized embedding (prompts) to each layer input of the Transformer. Such random initialization cannot match well to the point cloud. In addition, during the testing phase, the prompts are static, i.e., they cannot generate instance-specific representations for each point cloud.

To this end, we leverage the Dynamic Adapter to generate the prompt derived from the original model's internal output using the proposed Dynamic Adapter, called Internal Prompt tuning. Specifically, Dynamic Adapter operates on raw tokens, allowing our prompt to incorporate the model's pre-training knowledge effectively. In practice, we empirically average pool the output of Dynamic Adapter after applying a GELU activation. Then, we utilized TFTS to obtain a Prompt and concatenated it between the classification token $\boldsymbol{T}_{cls}\in \mathbb{R}^{1 \times d}$ and patch tokens $\boldsymbol{T}\in \mathbb{R}^{N \times d}$. The output of the $i$-th transformer layer $\boldsymbol{x}_{i}$ can be expressed as
\begin{equation}
\boldsymbol{x}_{i} = L _i\left (\left [ \boldsymbol{T} _{cls}; \boldsymbol{P}_{0},\ldots , \boldsymbol{P} _{i-1}; \boldsymbol{T}\right ] \right ).
\end{equation}

Our Dynamic Adapter adds a residual embedding ($\Delta\boldsymbol{x}_k $) for each token ($\boldsymbol{x}_k$), acting as a local perspective without modeling other point tokens (i.e., $\Delta \boldsymbol{x}_k$ only comes from $\boldsymbol{x}_k$). By pooling all outputs of the Dynamic Adapter, our internal prompts ($\textstyle \sum\Delta\boldsymbol{x}_k $) can serve as a global perspective and interact directly with each point tokens $\boldsymbol{x}_k+\Delta\boldsymbol{x}_k$, enabling access to the cumulative information of the entire sequence. Besides, the bi-directional interaction in self-attention strengthens the representation of the Dynamic Adapter, further improving the performance.

Please refer to Algorithm~\ref{alg:DA} for more detailed information on Dynamic Adapter and Internal Prompt.

\begin{algorithm}[t]
\small
\caption{\small Dynamic Adapter and Internal Prompt.
}
\label{alg:DA}
\definecolor{codeblue}{rgb}{0.25,0.5,0.5}
\definecolor{codekw}{rgb}{0.85, 0.18, 0.50}
\lstset{
  backgroundcolor=\color{white},
  basicstyle=\fontsize{7.5pt}{7.5pt}\ttfamily\selectfont,
  columns=fullflexible,
  breaklines=true,
  captionpos=b,
  commentstyle=\fontsize{7.5pt}{7.5pt}\color{codeblue},
  keywordstyle=\fontsize{7.5pt}{7.5pt}\color{codekw},
  escapechar={|}, 
}
\begin{lstlisting}[language=python]
# An example of Dynamic Adapter and Internal Prompt
import torch.nn as nn
import torch.nn.functional as F

class DynamicAdapter():
    def __init__(self, dim, r):
        super().__init__()
        self.norm = nn.LayerNorm(dim)
        self.scale = nn.Linear(dim, 1)
        self.down_proj = nn.Linear(dim, r)
        self.up_proj = nn.Linear(r, dim)
        self.active = nn.GELU()
        with torch.no_grad():
            nn.init.zeros_(self.up_proj.weight)
            nn.init.zeros_(self.up_proj.bias)
    def forward(self, x):
        # Dynamic Adapter
        x = self.norm(x)
        dynamic_scale = F.relu(self.scale(x))
        out = self.down_proj(x)
        out = self.active(out)
        out = self.up_proj(out)
        out = out * dynamic_scale
        # Generate Internal Prompt
        internal_prompt = self.active(out).mean(dim=1)
        return out, internal_prompt
\end{lstlisting}
\vspace{-5pt}
\end{algorithm}
\label{alg:dynamic}

\subsection{Analysis}

\textbf{Tunable parameters.} The proposed DAPT keeps the entire backbone frozen and only tunes the newly inserted parameters, effectively reducing the number of tunable parameters. Specifically, the TFTS introduces only two parameters in $\mathbb{R}^{1 \times d}$ for each operation. Each extra Dynamic Adapter module adds $2d$ from layer norm, $d$ from $\boldsymbol{W}_{s}$, $2\times r\times d$ from $\boldsymbol{W}_{d}$ and $\boldsymbol{W}_{u}$ (bias ignored) as tunable parameters, where the intermediate dimension $r$ is relatively small compared to $d$. Our Internal Prompt tuning can propose a prompt by pooling and TFTS without introducing other parameters. In summary, we only need to fine-tune fewer than 5\% of the parameters to achieve comparable results.

\textbf{Training memory.} The proposed approach can consistently reduce training memory as the batch size increases. As shown in Fig.~\ref{fig:pipeline}(c), with 512 batch size, our DAPT significantly reduces GPU memory\footnote{The gradient checkpoint~\cite{chen2016training} is already being used to save memory.}
% \textsuperscript{\textcolor{red}{1}}
usage by 35\% and 49\% compared to full fine-tuning and IDPT~\cite{zha2023instance} (the recent SOTA PETL method for point cloud), respectively. 

\textbf{Storage.} The proposed DAPT significantly reduces the storage cost. After being fine-tuned on $M$ point cloud datasets, only one copy of the original pre-trained model’s full parameters and $M$ dataset-specific trainable parameters need to be stored, saving up to 95\% of storage space. Thus, even if storage is not a current concern, DAPT presents a promising approach for adapting the expansion of increasingly larger models in 3D vision.

\begin{table*}[ht]
    \footnotesize
    \setlength{\tabcolsep}{0.99mm}
    \centering

  \caption{
Classification on three variants of the ScanObjectNN~\cite{uy2019revisiting} and the ModelNet40~\cite{wu20153d}, including the number of trainable parameters and overall accuracy (OA). All methods utilize the default data argumentation as the baseline. \textcolor{red}{$^*$} denotes reproduced results. We report ScanObjectNN results without voting. ModelNet40 results are without and with voting, referred to (-/-).
}
  \vspace{-10pt}
    \begin{tabular}{lcccccccc}
    
    \toprule
    \multirow{2.3}{*}{Method} &\multirow{2.3}{*}{Reference} &\multirow{2.3}{*}{Tunable params. (M)} &\multirow{2.3}{*}{FLOPs (G)} &\multicolumn{3}{c}{ScanObjectNN} &\multicolumn{2}{c}{ModelNet40}\\
		\cmidrule(r){5-7} \cmidrule{8-9}
	& & & &OBJ\_BG & OBJ\_ONLY &PB\_T50\_RS & Points Num. & OA (\%)      \\
    \midrule
    \multicolumn{9}{c}{\textit{Supervised Learning Only}} \\
    \midrule
    PointNet~\cite{qi2017pointnet} & CVPR 17 & 3.5 & 0.5  & 73.3  & 79.2  & 68.0 & 1k & - / 89.2 \\
    PointNet++~\cite{qi2017pointnet++}   & NeurIPS 17 & 1.5 & 1.7 & 82.3  & 84.3  & 77.9 & 1k & - / 90.7\\
    DGCNN~\cite{wang2019dynamic}  & TOG 19 & 1.8 & 2.4 & 82.8  & 86.2  & 78.1 & 1k & - / 92.9 \\
    MVTN~\cite{hamdi2021mvtn}  & ICCV 21 & 11.2 & 43.7 & -     & -     & 82.8 & 1k & - / 93.8\\
    PointNeXt~\cite{qian2022pointnext}  & NeurIPS 22  & 1.4 & 1.6 & -     & -  & 87.7 & 1k & - / 94.0\\
    PointMLP~\cite{ma2022rethinking}  & ICLR 22 &  13.2 & 31.4  & -    & -     & 85.4  & 1k & - / 94.5\\
    RepSurf-U~\cite{ran2022surface} & CVPR 22 & 1.5   & 0.8 &  -  & -    & 84.3  & 1k  & - / 94.4 \\
    ADS~\cite{hong2023attention} & ICCV 23 & -  & -  &  - & -   & 87.5 & 1k  & - / 95.1 \\
    \midrule
    \multicolumn{9}{c}{\textit{ Self-Supervised Representation Learning (Full fine-tuning)}} \\
    \midrule
    % Transformer~\cite{} &  & 22.1 & & 79.86 & 80.55 &  77.24 & 1k & 91.4 \\
    OcCo~\cite{wang2021unsupervised} & ICCV 21 & 22.1 & 4.8 & 84.85 & 85.54 & 78.79 & 1k & - / 92.1 \\
    Point-BERT~\cite{yu2022point}  & CVPR 22 & 22.1 & 4.8  & 87.43 & 88.12 &  83.07 & 1k & - / 93.2 \\
    MaskPoint~\cite{liu2022masked} & ECCV 22 & 22.1 & - & 89.70 & 89.30 &  84.60 & 1k & - / 93.8 \\
    Point-MAE~\cite{pang2022masked}  & ECCV 22 & 22.1 & 4.8 & 90.02 & 88.29 & 85.18 & 1k & - / 93.8 \\
    Point-M2AE~\cite{zhang2022point}  & NeurIPS 22 & 15.3 & 3.6 & 91.22 & 88.81 & 86.43 
 & 1k & - / 94.0\\
    ACT~\cite{dong2022autoencoders}   & ICLR 23 & 22.1 & 4.8 & 93.29 & 91.91  & 88.21 & 1k & - / 93.7\\
    \recon~\cite{qi2023recon} & ICML 23& 43.6 & 5.3 & 94.15 & 93.12  & 89.73 & 1k & - / 93.9  \\
    
    \midrule
    \multicolumn{9}{c}{\textit{Self-Supervised Representation Learning (Efficient fine-tuning)}} \\
    
    \midrule
    Point-BERT~\cite{yu2022point} (baseline)  & CVPR 22 & 22.1 (100\%) & 4.8 & 87.43 & 88.12 & 83.07& 1k & {92.7} / {\color{gray}{93.2}}\\
    + IDPT~\cite{zha2023instance}& ICCV 23 & 1.7 (7.69\%)& 7.2 & {88.12}\dplus{+0.69}  & {88.30}\dplus{+0.18}  & {83.69}\dplus{+0.62}  & 1k & 92.6{\dtplus{-0.1}} / {\color{gray}{{93.4}}}{\color{gray}{\ddplus{+0.2}}}\\
    \rowcolor{linecolor!40}+ DAPT ({ours})& - & \textbf{1.1} (\textbf{4.97}\%) & 5.0 & {91.05}\dplus{+3.62} & {89.67}\dplus{+1.55} & {85.43}\dplus{+2.36} &1k & {93.1}{\dplus{+0.4}} / {\color{gray}{{93.6}}}{\color{gray}{\ddplus{+0.4}}} \\
    \midrule
    % Point-MAE (baseline\textcolor{red}{$^*$})& 22.1  & 90.01 & 87.95 & 84.39 & 1k & 93.2 / 93.8\\
    Point-MAE~\cite{pang2022masked} (baseline)& ECCV 22 & 22.1 (100\%)& 4.8& 90.02 & 88.29 & {85.18} & 1k & 93.2 / {\color{gray}{93.8}}\\
    + IDPT~\cite{zha2023instance}& ICCV 23 & 1.7 (7.69\%) & 7.2 & {91.22}\dplus{+1.20} & {90.02}\dplus{+1.73}& 84.94\dtplus{-0.24} & 1k & {93.3}{\dplus{+0.1}} / {\color{gray}{{94.4}}}{\color{gray}{\ddplus{+0.6}}} \\
    \rowcolor{linecolor!40}+ DAPT ({ours})& - & \textbf{1.1} (\textbf{4.97}\%) & 5.0 & {90.88}\dplus{+0.86} & {90.19}\dplus{+1.90} & {85.08}\dtplus{-0.10} & 1k & {93.5}{\dplus{+0.3}} / {\color{gray}{{94.0}}}{\color{gray}{\ddplus{+0.2}}} \\
    \midrule
    \recon~\cite{qi2023recon} (baseline\tablefootnote{We modify the structure by loading a pre-trained \recon~weight into Point-MAE as a single-modal fine-tuning approach to reproduce similar results as the original paper.\label{recon_modify}})& ICML 23  & 22.1 (100\%)& 4.8 & {94.32} & {92.77} & {90.01}& 1k & 92.5 / {\color{gray}{93.0}}\\
+ IDPT\textcolor{red}{$^*$}~\cite{zha2023instance}& ICCV 23  & 1.7  (7.69\%) & 7.2 &93.29\dtplus{-1.03} &91.57\dtplus{-1.20} & 87.27\dtplus{-2.74} & 1k & {93.4}{\dplus{+0.9}} / {\color{gray}{{93.5}}}{\color{gray}{\ddplus{+0.5}}} \\ 
    \rowcolor{linecolor!40}+ DAPT ({ours})& -  & \textbf{1.1} (\textbf{4.97}\%) & 5.0 &{94.32}\ddplus{0.00} & {92.43}\dtplus{-0.34} & {89.38}\dtplus{-0.63} & 1k &  {93.5}{\dplus{+1.0}} / {\color{gray}{{94.1}}}{\color{gray}{\ddplus{+1.1}}}\\
    \bottomrule
    \end{tabular}%
  
      \label{tab:joint_tab}

\end{table*}%

\section{Experiments}
\subsection{Implement Details}
To ensure a fair comparison, we employed identical experimental settings to the default fine-tuning method for each baseline. This entails freezing the weights of the pre-trained point cloud backbone and solely updating the newly inserted parameters during training. All experiments are conducted on a single GeForce RTX 4090.

\subsection{3D Classification}

\textbf{Real-World Object Classification.} ScanObjectNN~\cite{uy2019revisiting} is a highly challenging 3D dataset covering $\sim$$15$K real-world objects across $15$ categories. These objects consist of indoor scene data obtained by scanning, often characterized by cluttered backgrounds and occlusion caused by other objects. As shown in Tab.~\ref{tab:joint_tab}, we conducted experiments on three variants of ScanObjectNN (i.e., OBJ\_BG, OBJ\_ONLY, and PB\_T50\_RS), each with increasing complexity. We utilize various baselines such as Point-BERT~\cite{yu2022point}, Point-MAE~\cite{pang2022masked}, and \recon~\cite{qi2023recon} to demonstrate the effectiveness and generalization of our approach. It is worth noting that our DAPT meets or exceeds the performance of full fine-tuning in most cases, while only tuning less than 5\% of the parameters. Especially, DAPT achieved increases of 3.62\%, 1.55\%, and 2.36\% in the three ScanObjectNN variants on Point-BERT. Furthermore, we surpass IDPT~\cite{zha2023instance}, the most recent SOTA method for 3D efficient fine-tuning, in almost every variant for different baselines, with 35\% fewer tunable parameters. A similar phenomenon is also observed in Point-MAE and \recon. Moreover, our DAPT has only a slight increase in FLOPs compared to the full fine-tuning and is quite competitive with IDPT.

\textbf{Synthetic Object Classification.} ModelNet40~\cite{wu20153d} contains $12,311$ pristine 3D CAD models across $40$ categories. Every point cloud is complete, uniform, and noise-free. Due to the voting strategy~\cite{liu2019relation} being time-consuming, we prioritize reporting overall Accuracy without voting. As shown in Tab.~\ref{tab:joint_tab}, although the performance in ModelNet40 is quite saturated, our method consistently achieves performance gains compared to different baselines and IDPT~\cite{zha2023instance}. Specifically, \recon~combined with our DAPT achieves 1.0\% improvement and significantly reduces tunable parameters. Even after voting, our DAPT remains the leader.

\textbf{Few-shot Learning.} We further conduct few-shot experiments on ModelNet40~\cite{wu20153d} to prove our few-shot transfer learning ability. Consistent with previous works~\cite{yu2022point,pang2022masked}, we employ the n-way, m-shot configuration. As shown in Tab.~\ref{tab:fewshot}, it can be observed that DAPT achieved performance enhancements in most cases compared to full fine-tuning and IDPT, indicating its effectiveness in few-shot learning.

\subsection{Part Segmentation}
Part segmentation is challenging to predict a more detailed class label for every point. We evaluate the effectiveness of DAPT on ShapeNetPart~\cite{yi2016scalable}, which includes $16,881$ samples from $16$ categories. Following previous works~\cite{pang2022masked,dong2022autoencoders}, we incorporate Dynamic Adapter into the $2$nd, $6$th, and $10$th layers, generating an Internal Prompt to feed into the input of the $3$rd, $7$th, and $11$th layers and employing them as global features. As shown in Tab.~\ref{tab:segmentation}, our DAPT achieves competitive results on Inst. mIoU and notable improvements on Cls. mIoU compared with IDPT~\cite{zha2023instance}. Keep in mind that the increase in tunable parameters mainly comes from the head part, but we still reduce the tunable parameters and gain improvements compared with IDPT.

\begin{table}[!t]
  \centering
  \scriptsize
    \setlength{\tabcolsep}{0.5mm}
  \caption{Few-shot learning on ModelNet40~\cite{wu20153d}. Overall accuracy (\%)$\pm$the standard deviation (\%) without voting is reported.}
  \vspace{-10pt}
    \begin{tabular}{lccccc}
    \toprule
   \multirow{2.3}{*}{Methods}&\multirow{2.3}{*}{Reference} & \multicolumn{2}{c}{5-way} & \multicolumn{2}{c}{10-way} \\
\cmidrule{3-6}  &        & 10-shot & 20-shot & 10-shot & 20-shot \\
    \midrule
    \multicolumn{6}{c}{\textit{with Self-Supervised Representation Learning (Full fine-tuning)}} \\
    \midrule
    OcCo~\cite{wang2021unsupervised}& ICCV 21      & 94.0$\pm$3.6& 95.9$\pm$2.3 & 89.4$\pm$5.1 & 92.4$\pm$4.6 \\
    Point-BERT~\cite{yu2022point}  &  CVPR 22    & 94.6$\pm$3.1 & 96.3$\pm$2.7 & 91.0$\pm$5.4 & 92.7$\pm$5.1 \\
    MaskPoint~\cite{liu2022masked}  &   ECCV 22   & 95.0$\pm$3.7 & 97.2$\pm$1.7 & 91.4$\pm$4.0 & 93.4$\pm$3.5 \\
    Point-MAE~\cite{pang2022masked} &   ECCV 22   & 96.3$\pm$2.5 & 97.8$\pm$1.8 & 92.6$\pm$4.1 & 95.0$\pm$3.0 \\
    Point-M2AE~\cite{zhang2022point} &  NeurIPS 22    & 96.8$\pm$1.8 & 98.3$\pm$1.4 & 92.3$\pm$4.5 & 95.0$\pm$3.0 \\
    ACT~\cite{dong2022autoencoders}  & ICLR 23      & 96.8$\pm$2.3 & 98.0$\pm$1.4 & 93.3$\pm$4.0 & 95.6$\pm$2.8 \\
    \recon~\cite{qi2023recon}  & ICML 23      & 97.3$\pm$1.9 & 98.9$\pm$3.9 & 93.3$\pm$3.9 & 95.8$\pm$3.0 \\
    \midrule
    \multicolumn{6}{c}{\textit{with Self-Supervised Representation Learning (Efficient fine-tuning)}} \\
    \midrule
   Point-BERT~\cite{yu2022point} (baseline)  & CVPR 22 &94.6$\pm$3.1 & 96.3$\pm$2.7 & 91.0$\pm$5.4 & 92.7$\pm$5.1 \\
   + IDPT~\cite{zha2023instance}  & ICCV 23    & \textbf{96.0}$\pm$\textbf{1.7}& 97.2$\pm$2.6& 91.9$\pm$4.4& 93.6$\pm$3.5\\
   \rowcolor{linecolor!40}+ DAPT (\textbf{ours}) & -&95.8$\pm$2.1 &\textbf{97.3}$\pm$\textbf{1.3} &\textbf{92.2}$\pm$\textbf{4.3} &\textbf{94.2}$\pm$\textbf{3.4} \\
       \midrule
    Point-MAE~~\cite{pang2022masked} (baseline)&ECCV 22   & 96.3$\pm$2.5 & 97.8$\pm$1.8 & 92.6$\pm$4.1 & 95.0$\pm$3.0\\
   + IDPT~\cite{zha2023instance} &   ICCV 23    & \textbf{97.3}$\pm$2.1& 97.9$\pm$1.1& 92.8$\pm$4.1& 95.4$\pm$\textbf{2.9}\\
   \rowcolor{linecolor!40}+ DAPT (\textbf{ours}) & - & 96.8$\pm$\textbf{1.8}  &  \textbf{98.0}$\pm$\textbf{1.0} & \textbf{93.0}$\pm$\textbf{3.5} & \textbf{95.5}$\pm$3.2  \\
    \bottomrule
    \end{tabular}%
  
  \label{tab:fewshot}%
\end{table}%

\begin{table}[t]
  \centering
  \scriptsize
  \setlength{\tabcolsep}{0.6mm}
  \caption{Part segmentation on the ShapeNetPart~\cite{yi2016scalable}. The mIoU for all classes (Cls.) and for all instances (Inst.) are reported. \#TP represents the tunable parameters. \textcolor{red}{$^*$} denotes reproduced results.}
    \vspace{-10pt}
    \begin{tabular}{lcccc}
    \toprule
    Methods & Reference & \#TP (M)& Cls. mIoU (\%) & Inst. mIoU (\%) \\
    \midrule
    \multicolumn{5}{c}{\textit{Supervised Learning Only}} \\
    \midrule
    PointNet \cite{qi2017pointnet} & CVPR 17  &- & 80.39 & 83.7 \\
    PointNet++  \cite{qi2017pointnet++}  & NeurIPS 17 &-  & 81.85 & 85.1 \\
    DGCNN \cite{wang2019dynamic} & TOG 19 & - & 82.33 & 85.2 \\
    APES~\cite{wu2023attention} & CVPR 23& - & 83.67 & 85.8\\
    \midrule
    \multicolumn{5}{c}{\textit{ Self-Supervised Representation Learning (Full fine-tuning)}} \\
    \midrule
    % Transformer \cite{} & & 27.09 & 83.42 & 85.1 \\
    OcCo \cite{wang2021unsupervised} & ICCV 21 & 27.09 & 83.42 & 85.1 \\
    MaskPoint \cite{liu2022masked} & ECCV 22 & - & 84.60 & 86.0 \\
    Point-BERT \cite{yu2022point} & CVPR 22 & 27.09 & 84.11 & 85.6 \\
    Point-MAE \cite{pang2022masked} & ECCV 22 & 27.06 & 84.19 & 86.1 \\ 
    ACT \cite{dong2022autoencoders} & ICLR 23 &  27.06 & 84.66 & 86.1 \\
    \midrule
    \multicolumn{5}{c}{\textit{ Self-Supervised Representation Learning (Efficient fine-tuning)}} \\
    \midrule
    Point-BERT~\cite{yu2022point} (baseline) &  CVPR 22 & 27.06 & 84.11 & 85.6 \\ 
    + IDPT\textcolor{red}{$^*$}~\cite{zha2023instance} & ICCV 23 & 5.69  & 83.50  & 85.3  \\
    \rowcolor{linecolor!40}+ DAPT (\textbf{ours})& - & \textbf{5.65}  & 83.83 & 85.5 \\
    \midrule
    Point-MAE~\cite{pang2022masked} (baseline) &  ECCV 22 & 27.06 & 84.19 & 86.1 \\ 
    + IDPT~\cite{zha2023instance} & ICCV 23 & 5.69  & 83.79  & 85.7  \\
    \rowcolor{linecolor!40}+ DAPT (\textbf{ours})& - & \textbf{5.65}  & 84.01 & 85.7 \\
    \midrule
    \recon~\cite{qi2023recon}~(baseline\textsuperscript{\ref{recon_modify}}) & ICML 23 & 27.06 & 84.52 & 86.1 \\
    + IDPT\textcolor{red}{$^*$}~\cite{zha2023instance} &  ICCV 23 & 5.69 & 83.66  & 85.7 \\
    \rowcolor{linecolor!40}+ DAPT (\textbf{ours}) &- &\textbf{5.65}  &83.87 & 85.7\\
    \bottomrule
    \end{tabular}
    \vspace{-10pt}
  \label{tab:segmentation}
\end{table}

\subsection{Compared with Other PETL Methods}

\begin{table}[!t]
\scriptsize
\setlength{\tabcolsep}{3.99mm}
\centering
\caption{Comparisons of parameter efficient transfer learning methods from NLP and 2D Vision on the hardest variant of ScanObjectNN~\cite{uy2019revisiting}. Overall accuracy (\%) without voting is reported. \#TP represents the tunable parameters.}
  \vspace{-10pt}
\label{tab:nlp_compare}
\begin{tabular}{ lccc }
\toprule
 Method &Reference& \#TP (M) & PB\_T50\_RS \\
\midrule
 Point-MAE~\cite{liu2022masked}  &ECCV 22 & 22.1 & \textbf{85.18}  \\
 Linear probing &- & 0.3& 75.99\\
 \midrule
  + Adapter~\cite{houlsby2019parameter}&ICML 19 & 0.9 & 83.93 \\
  + Perfix tuning~\cite{li2021prefix}& ACL 21 &0.7 & 77.72  \\
  + BitFit~\cite{zaken2022bitfit} & ACL 21 &0.3 & 82.62    \\
  + LoRA~\cite{hu2021lora} & ICLR 22 & 0.9&  81.74   \\
  + VPT-Deep~\cite{jia2022visual}&ECCV 22 &0.4 &  81.09 \\
  + AdaptFormer~\cite{chen2022adaptformer} &NeurIPS 22 &0.9  & 83.45 \\
  + SSF~\cite{lian2022scaling} & NeurIPS 22  &0.4  & 82.58\\
  \midrule
  + IDPT~\cite{zha2023instance} &ICCV 23 & 1.7 &84.94\\
  \rowcolor{linecolor!40}+ DAPT (\textbf{ours}) & - & 1.1 & \textbf{85.08} \\
\bottomrule
\end{tabular}
\end{table}

\begin{table}[!t]
\scriptsize
\setlength{\tabcolsep}{2.63mm}
\centering
 % \footnotesize
\caption{The effect of each component of our DAPT. The tunable parameters (\#TP) and the overall accuracy (\%) on the hardest variant of ScanObjectNN~\cite{uy2019revisiting} are reported. In. and Ex. indicate Internal and External respectively.}
  \vspace{-10pt}
\label{tab:compoments}

\begin{tabular}{ cccccc }
   \toprule
 TFTS & DA & In. Prompt & Ex. Prompt &  \#TP (M) & PB\_T50\_RS \\
\midrule
\multicolumn{4}{c}{Full fine-tuning}  &22.1 & 85.18 \\
\multicolumn{4}{c}{Linear Probing}  &0.27 & 75.99 \\
\midrule
\ding {52} & &  & & 0.37 &83.17 \\
&\ding {52} &  & & 0.88 &84.25 \\
\ding {52}&\ding {52} &  & & 0.99 & 84.80\\
\ding {52}&\ding {52} &  &\ding {52} & 1.08& 84.70\\
\rowcolor{linecolor!40}\ding {52}  &\ding {52} & \ding {52}&   & 1.09& \textbf{85.08} \\

\bottomrule
\end{tabular}
\vspace{-15pt}
\end{table}

To further prove the effectiveness of our proposed method, we compare DAPT with several PETL approaches from NLP and 2D vision. As shown in Tab.~\ref{tab:nlp_compare}, we select the most challenging PB\_T50\_RS variant with Point-MAE as the baseline. Note that we have carefully tuned the hyperparameters of these methods to attain optimal performance. While they show different performance improvements compared to linear probing, there is still a considerable gap compared to full fine-tuning. Although these methods excel in NLP or 2D, they still struggle to adapt to 3D data due to missing point clouds and other complexities. It is worth noting that our method surpasses the IDPT while reducing the number of tunable parameters by 35\% and achieving a better trade-off between tunable parameters and performance.

\subsection{Ablation Study}
We conducted ablation studies on the most challenging PB\_T50\_RS variant to investigate the rationalization and effectiveness of the design of our proposed DAPT and then fine-tuned on Point-MAE as the baseline.

\begin{table*}
    \centering
    \scriptsize
    \caption{Ablation on hyper-parameters and settings of DAPT, including dimension $r$, settings of dynamic scale and inserted layers. The tunable parameters (\#TP) and the overall accuracy (\%) on the hardest variant of ScanObjectNN~\cite{uy2019revisiting} are reported.}
      \vspace{-10pt}
    \resizebox{0.95\linewidth}{!}{
    \begin{subtable}[htbp]{0.25\linewidth} % 0.245
        \centering
        \scriptsize
         \setlength{\tabcolsep}{1.2mm} % 1.2mm
        \caption{Ablation on $r$ of Dynamic Adapter.}
        \label{tab:rank}
        \begin{tabular}{ccc}
            \toprule
            Dimension $r$& \#TP (M) & PB\_T50\_RS \\
            \midrule
            8 & 0.57 & 84.39  \\
            16 &0.64  & 84.52 \\
            32 & 0.79 & 84.18 \\
           \rowcolor{linecolor!40} 64 & 1.09 & \textbf{85.08} \\
            72 & 1.16 &84.46 \\
            \bottomrule
        \end{tabular}
    \end{subtable}

    \begin{subtable}[htbp]{0.4\linewidth} % 0.235
        \centering
        \scriptsize
         \setlength{\tabcolsep}{1.2mm} % 0.2mm
        \caption{Ablation on scale $S_d$ in Dynamic Adapter.}
        \label{tab:scale}
        \begin{tabular}{ccc}
            \toprule
             Scale &\#TP (M) &PB\_T50\_RS \\
            \midrule
              1.0 & 1.08 &  84.59 \\
              5.0 & 1.08 &  84.42 \\
              10.0& 1.08 & 84.11  \\
              Dynamic scale w/o ReLU  &1.09  &84.73 \\
              \rowcolor{linecolor!40}Dynamic scale  &1.09  &\textbf{85.08} \\
            \bottomrule
        \end{tabular}
    \end{subtable}

    \begin{subtable}[htbp]{0.22\linewidth} % 0.22
    \setlength{\tabcolsep}{1.9mm} % 1.3mm
        \centering
         \scriptsize
        \caption{Ablation on the inserted layers.}
        \label{tab:insert}
     
        \begin{tabularx}{\textwidth}{ccc}
            \toprule
            Layers &  \#TP (M)  & PB\_T50\_RS \\
            \midrule
            1$\rightarrow$3  & 0.62 &80.67 \\
            1$\rightarrow$6  & 0.78 &82.48\\
            1$\rightarrow$9  & 0.93 &83.97\\
            4$\rightarrow$12 & 0.93 &81.75\\
            \rowcolor{linecolor!40} 1$\rightarrow$12  &1.09 & \textbf{85.08}\\
           
            \bottomrule 
        \end{tabularx}
    \end{subtable}

}
\end{table*}

\textbf{Analysis on each component.} We conduct experiments to prove the effectiveness of the proposed components of our DAPT and display the results in Tab.~\ref{tab:compoments}. When no additional inserted components exist, the method is the same as Linear Probing, which only achieves 75.99\% overall accuracy. By adopting our TFTS, we can achieve a substantial 7.18\% performance increase while only adding 0.1M tunable parameters. To examine the effectiveness of our proposed Dynamic Adapter, we exclusively use Dynamic Adapter and achieve an 8.26\% performance gain. These results indicate that TFTS and Dynamic Adapter are powerful parameter-efficient transfer learning modules. We further combine the above two components to gain an 84.80\% performance. We also externally replaced the Internal Prompt, using a Xavier uniform initializer~\cite{jia2022visual} mentioned in VPT~\cite{jia2022visual}. Compared to our Internal Prompt, the External Prompt shows a 0.38\% performance fall and even a 0.1\% performance drop compared with no Prompt tuning, demonstrating the effectiveness of Internal Prompts.

\begin{figure}[t]
	\begin{center}
		\includegraphics[width=0.99\linewidth]{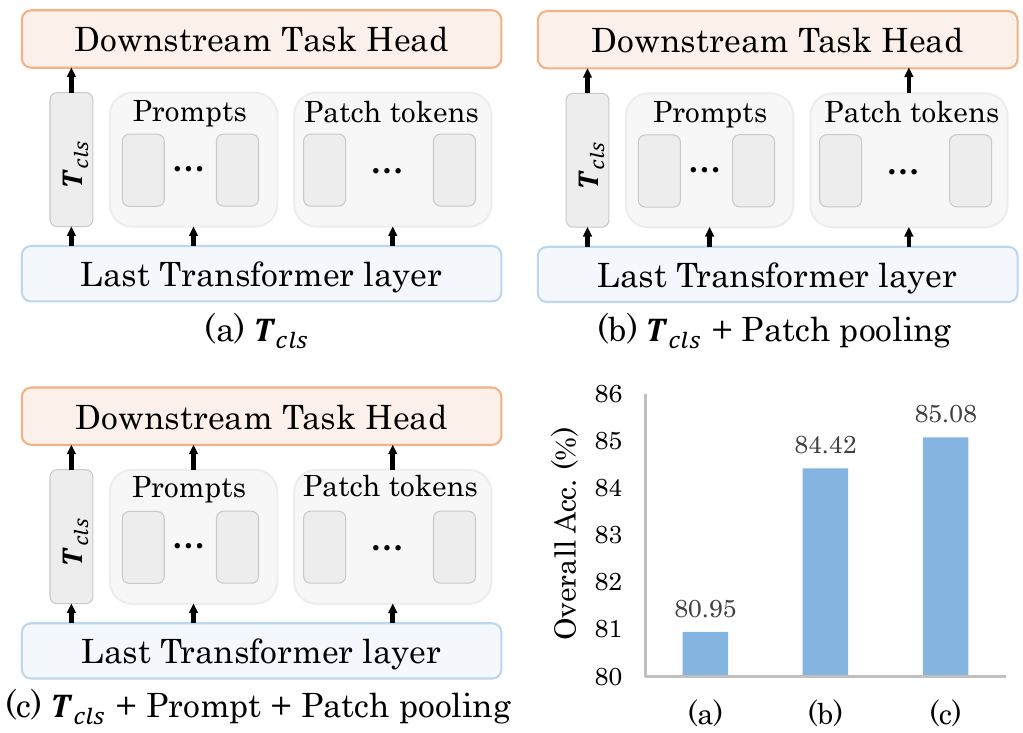}
	\end{center}
    \vspace{-16pt}
	\caption{Effect of different inputs for downstream task head. We conduct experiments on the hardest variant (i.e., PB\_T50\_RS) of ScanObjectNN~\cite{uy2019revisiting} with Point-MAE~\cite{pang2022masked} baseline.}
	\label{fig:head}
     \vspace{-8pt}
\end{figure}

\textbf{Analysis on token selections for head inputs.} We examined the influence of the input tokens of the downstream task head, including the classification token, the pooling of our prompt tokens, and the pooling of point patch tokens. As depicted in Fig.~\ref{fig:head}, we selected three standard token selection methods. We observe that the best performance is attained when all three features are included: $\boldsymbol{T}_{cls}$, prompt tokens, and point patch tokens. Comparing the use of $\boldsymbol{T}_{cls}$ only and both $\boldsymbol{T}_{cls}$ and pooling of Patch tokens, using all the features resulted in a 4.13\% and 0.66\% performance gain.

\textbf{Analysis on dimension $r$ in Dynamic Adapter.} One of our keys to reducing tunable parameters is selecting dimension $r$ in Dynamic Adapter. As shown in Tab.~\ref{tab:rank}, we conduct experiments on various values of $r$ and find that when $r=64$, we can achieve the best trade-off between tunable parameters and performance.

\begin{figure}[!t]
	\begin{center}
		\includegraphics[width=0.96\linewidth]{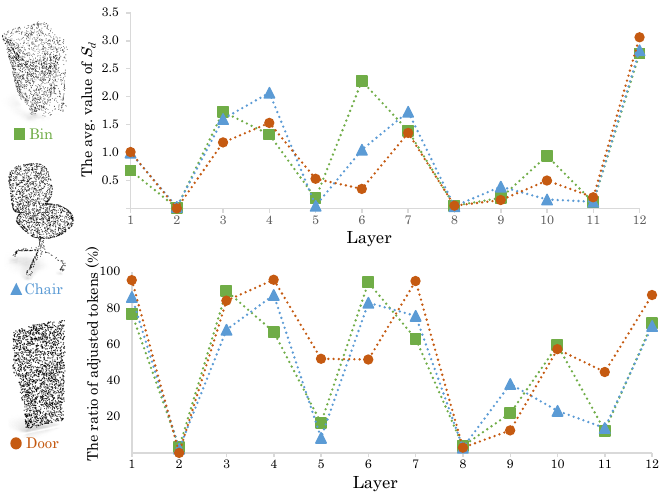}
	\end{center}
    \vspace{-18pt}
	\caption{A visual results of the average value of $S_d$ and the ratio of adjusted tokens for each layer across three samples in OBJ\_ONLY variant of ScanObjectNN~\cite{uy2019revisiting}.}
     \vspace{-6pt}
	\label{fig:scale}
\end{figure}

\textbf{Analysis on scale $S_d$ in Dynamic Adapter.} It is crucial to obtain a dynamic scale during inference due to the complexity of point clouds. As indicated in Tab.~\ref{tab:scale}, fixing the scale to $1.0$ yields the best result but decreases by 0.49\% compared to our dynamic scale. We also perform a variation of dynamic scale, removing the ReLU, and observe a 0.35\% decrease, which is still superior to the fixed scale. In addition, we displayed the average value of $S_d$ and the ratio of adjusted tokens in each layer by choosing three samples in Fig.~\ref{fig:scale}. While their trends are similar, various samples differ across different layers, demonstrating that our method can dynamically generate scale for different samples to overcome the complex geometry structures of point clouds.

\textbf{Analysis on inserted layers for Dynamic Adapter and Internal Prompts.} One way to further reduce the tunable parameters is to select only a portion of the layer to insert our components. However, as shown in Tab.~\ref{tab:insert}, we observe that fine-tuning only shallow blocks and fine-tuning only deep blocks leads to different performance decline levels. The potential reason may be the difficulty in determining which layer is more important when the network processes the diverse point cloud. To address this issue, we incorporate our proposed components for each layer and utilize the dynamic scale to regulate the extent of feature adjustment.

\section{Conclusion and Limitation}
In this study, we propose a simple yet effective parameter-efficient transfer learning strategy named DAPT for point cloud analysis. The proposed DAPT freezes the parameters of the pre-trained backbones and leverages the Dynamic Adapter to adjust each token dynamically by considering the significance. In addition, we further use Dynamic Adapter to implement Internal Prompt tuning, better capturing the instance-specific feature. Our method provides a practical solution to reduce storage cost requirements without compromising performance. 
One limitation is that whether our method could perform well in more complex tasks such as 3D object detection and generation is still unclear, which is our future work.

\textbf{Acknowledgement.} This work was supported in part by the NSFC (No.62225603), in part by the Hubei Key R\&D Program (No.2022BAA078), in part by the Taihu Lake Innovation Fund for Future Technology (HUST:2023-A-1), and in part by the National Undergraduate Training Projects for Innovation and Entrepreneurship (202310487020).

%%%%%%%%% REFERENCES
{\small
\bibliographystyle{ieeenat_fullname}
\balance
\bibliography{main}
}

\clearpage
\nobalance
\appendix

\section{Additional Experiments}

\subsection{Training Detail}
We adopt downstream fine-tuning configuration following pioneer work Point-MAE~\cite{pang2022masked}. More details are provided in Tab.~\ref{tab:paramas}. Taking fine-tuning on ScanObjectNN~\cite{uy2019revisiting} as an example, the overall training includes 300 epochs, with a cosine learning rate~\cite{loshchilov2017sgdr} of 5e-4, and a 10-epoch warm-up period. AdamW optimizer~\cite{loshchilov2019decoupled} is used.

\begin{table*}[t]
\footnotesize
\setlength{\tabcolsep}{8.0mm}
\captionof{table}{Training details for downstream fine-tuning.}
\vspace{-5pt}
\begin{tabular}{lccccc}
\toprule
    \multirow{2.3}{*}{Configuration}  &\multicolumn{3}{c}{Classification} & \multicolumn{2}{c}{Segmentation}\\
		\cmidrule(r){2-4} \cmidrule(r){5-6}
	 &ScanObjectNN & ModelNet & ModelNet Few-shot & ShapeNetPart     \\
    \midrule
 Optimizer & AdamW & AdamW & AdamW & AdamW \\
 Learning rate & 5e-4 & 5e-4 & 5e-4  & 2e-4\\
 Weight decay & 5e-2 & 5e-2 & 5e-2 & 5e-2 \\
 Learning rate scheduler & cosine & cosine & cosine & cosine \\
 Training epochs  & 300 & 300 & 150 & 300 \\
 Warmup epochs& 10 & 10& 10 & 10 \\
 Batch size & 32 & 32& 32 & 16 \\
 $r$ in Dynamic Adapter & 64 & 72 & 72 &128 \\
 \midrule
 Number of points  & 2048 & 1024& 1024 & 2048 \\
 Number of point patches & 128 & 64 & 64 & 128\\
 Point patch size  & 32 & 32 & 32  & 32 \\
\bottomrule
\vspace{10pt}
\end{tabular}
\label{tab:paramas}
\end{table*}

\subsection{Scale in Dynamic Adapter}

Based on Tab. 7(b) in the manuscript, we conduct additional experiments in Tab.~\ref{tab:appendix_scale} to further prove the effectiveness of our dynamic scale. Using the scale 0.1 suggested in AdaptFormer~\cite{chen2022adaptformer} or 4.0 suggested by He et al.~\cite{he2021towards} cannot achieve satisfying results. We also experiment with a learned scale which does not give better results. We claim that our dynamic scale offers greater adaptability for the intricate geometry of point clouds and eliminates the need to adjust scale as a hyper-parameter.

\subsection{Number of Internal Prompts}
In previous works~\cite{lester2021power,li2021prefix,jia2022visual}, external prompts are utilized by concatenating a certain number of adjustable tokens into the transformer's input space. Therefore, this subsection investigates the impact of prompt numbers in DAPT on classification tasks.

\begin{figure}[!h]
	\begin{center}
		\includegraphics[width=0.85\linewidth]{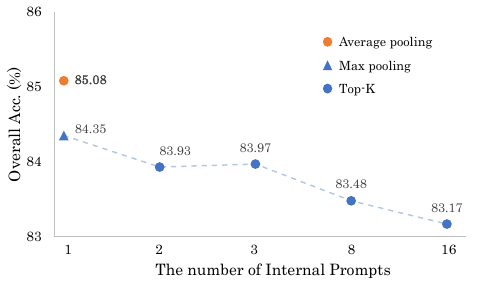}
	\end{center}
    \vspace{-15pt}
	\caption{The effect of the number of Internal Prompts.
 }
	\label{fig:num_prompt}
\end{figure}

We adopt average pooling (default), max pooling, and top-K operation to obtain internal prompts in different length. Fig.~\ref{fig:num_prompt} displays the results on the challenging variants (i.e., PE\_T50\_RS) of ScanObjectNN. The results suggest that simply increasing the prompt number may even hurt the performance.

\begin{table}[t]
\footnotesize
\setlength{\tabcolsep}{1.7mm}
\centering
\caption{The effect of different scale settings in Dynamic Adapter.}
\vspace{-10pt}
\label{tab:appendix_scale}
\begin{tabular}{cccc}
            \toprule
             Type& Scale &\#TP (M) &PB\_T50\_RS \\
            \midrule
            \multirow{2}{*}{Train: fixed}  &0.01& 1.08 &  83.28 \\
             \multirow{2}{*}{Inference: fixed} &0.1& 1.08 &  83.55 \\
              &4.0& 1.08 &  \textbf{84.35} \\
              % 1.0 & 1.08 &  84.59 \\
              % 5.0 & 1.08 &  84.42 \\
              % 10.0& 1.08 & 84.11  \\
            \midrule
            \multirow{5}{*}{Train: learnable}  &Initialized with 0.01 & 1.08 & 84.07\\
             \multirow{5}{*}{Inference: fixed} &Initialized with 0.1 & 1.08 & \textbf{84.77}\\
              &Initialized with 1.0 & 1.08 & 84.32\\
              & Initialized with 4.0 & 1.08 & 84.56\\
              & Initialized with 5.0 & 1.08 & 84.49\\
              & Initialized with 10.0& 1.08 & 84.70\\
             \midrule
              % Dynamic scale w/o ReLU  &1.09  &84.73 \\
              \multirow{0.33}{*}{Train: learnable}& \multirow{3}{*}{Dynamic scale}  &\multirow{3}{*}{1.09}  &\multirow{3}{*}{\textbf{85.08}} \\
               \multirow{0.33}{*}{Inference: dynamic}& & & \\
               \multirow{0.33}{*}{\textbf{(Ours)}}& & & \\

            \bottomrule
\end{tabular}
\end{table}
\begin{table}[!t]
\footnotesize
\setlength{\tabcolsep}{0.5mm}
\centering
\caption{Comparison of different fine-tuning strategy on ScanObjectNN classification. Throughput is measured with a batch size of 32 on a single RTX 4090 GPU.}
\vspace{-10pt}

\label{tab:inference_time}
\begin{tabular}{ lcccc }
\toprule
 Method & \#TP(M) & FLOPs(G) & Throughput (frame / s) & PB\_T50\_RS \\
 
\midrule
 Point-MAE~\cite{pang2022masked} & 22.1 & 4.8 & 323.66  & 85.18 \\
  IDPT~\cite{zha2023instance}  &  1.7 & 7.2 & 281.18 & 84.94 \\
  DAPT (Ours)  & 1.1 & 5.0 & 311.28 & 85.08 \\

\bottomrule
\vspace{-15pt}
\end{tabular}
\end{table}
\begin{figure*}[!t]
	\begin{center}
		\includegraphics[width=0.99\linewidth]{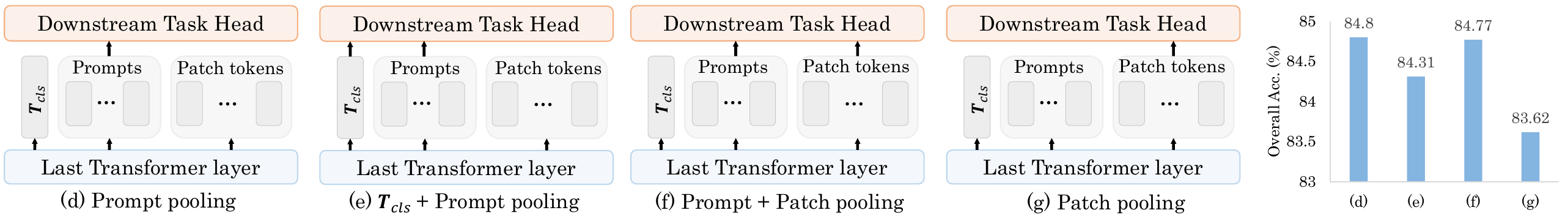}
	\end{center}
    \vspace{-10pt}
	\caption{The effect of different inputs for downstream task head. 
 }
	\label{fig:appendix_head}
     \vspace{-7pt}
\end{figure*}

\subsection{Token Selections for Head Inputs}
Based on Fig. 3 in the manuscript, we conduct four other experiments on token selections for head input, as shown in Fig.~\ref{fig:appendix_head}. Interestingly, with the pooling of our prompts, we can achieve better results than the pooling of patch tokens. The only use of the pooling of Prompts exceeds only the pooling of patch tokens by 1.18\%. We argue that our internal prompts better capture instance-specific features and act as global features, providing positive assistance to downstream task heads.

\subsection{Inference Time}
In this subsection we evaluate the inference time on classification task in Tab.~\ref{tab:inference_time}. Our DAPT achieves 311.28 frame/s with only a negligible impact on inference speed, which is highly competitive compared to IDPT (281.18 frame/s).

\section{Qualitative Analysis}
\subsection{t-SNE Visualizations}
\begin{figure*}[!t]
	\begin{center}
		\includegraphics[width=0.9\linewidth]{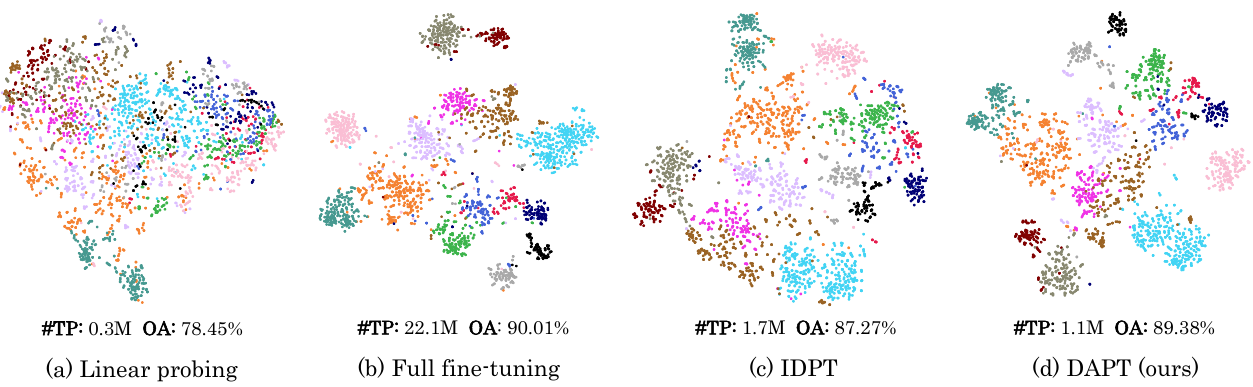}
	\end{center}
    \vspace{-10pt}
	\caption{The t-SNE visualizations from the test sets of ScanObjectNN (PB\_T50\_RS) using a pre-trained \recon~with different tuning strategies. We extract the final classification features from the top linear layer for t-SNE visualizations.}
	\label{fig:tsne}
     \vspace{-7pt}
\end{figure*}

In Fig.~\ref{fig:tsne}, the t-SNE~\cite{van2008visualizing} feature manifold visualization displays the models following full fine-tuning, linear probing, IDPT, and our DAPT on the ScanObjectNN PB\_T50\_RS dataset. From Fig.~\ref{fig:tsne}(a), it is evident that the feature distribution extracted by the ShapeNet-pretrained model on ScanObjectNN appears less discriminative. We contend that this is mainly due to the significant domain gap between the synthetic ShapeNet and real-world ScanObjectNN datasets, demonstrating the necessity for fine-tuning on downstream tasks. With full fine-tuning in Fig.~\ref{fig:tsne}(b), the feature distribution becomes more discriminative as all parameters are tuned. Fig.~\ref{fig:tsne}(c-d) confirms that our DAPT helps the pre-trained model generate more distinguishable representations with fewer learnable parameters than IDPT.

\begin{figure*}[t]
	\begin{center}
		\includegraphics[width=0.8\linewidth]{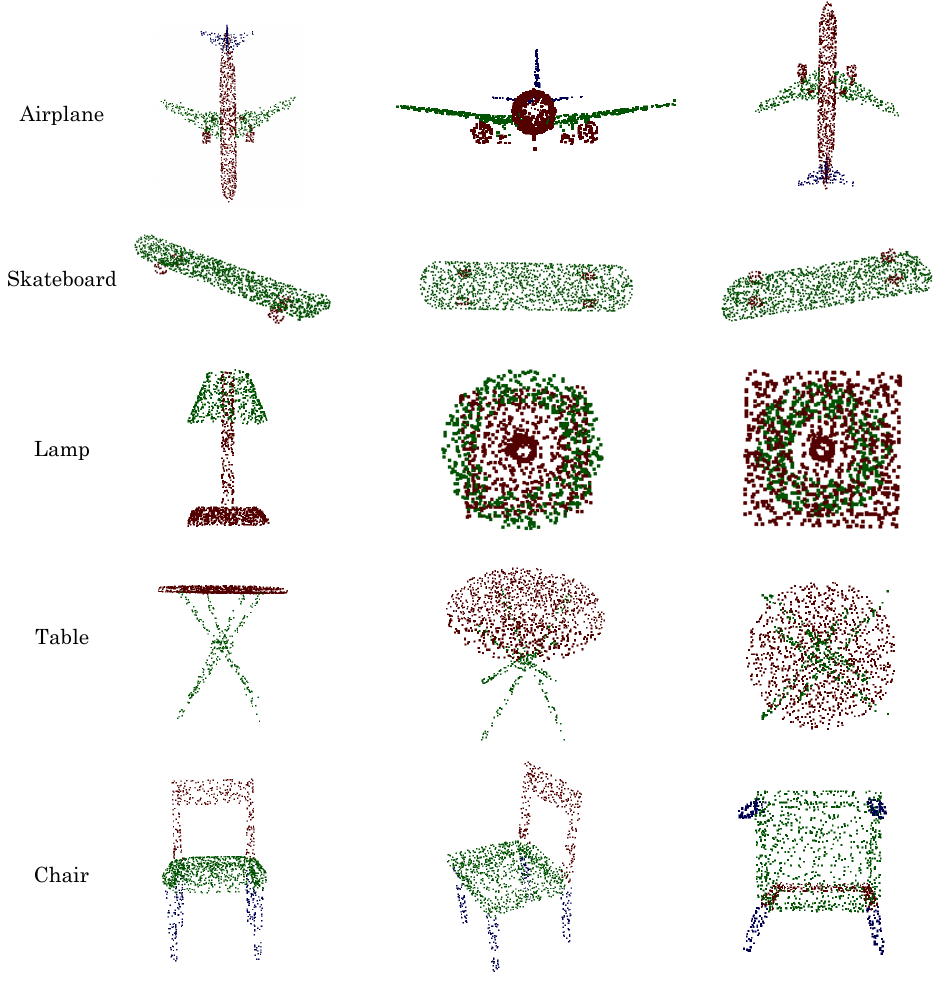}
	\end{center}
    \vspace{-15pt}
	\caption{Qualitative results for part segmentation. We show our prediction projection images from three different viewpoints.
 }
	\label{fig:part_vis}
\end{figure*}

\subsection{Part Segmentation Visualizations}
In Fig.~\ref{fig:part_vis}, we visualize our DAPT part segmentation results on Point-BERT~\cite{yu2022point} baseline. We select five representative categories each, each with three viewpoints. Our DAPT requires a small number of tunable parameters while achieving satisfying segment results.

\end{document}